\documentclass[runningheads]{llncs}
\usepackage{epsfig}
\usepackage{graphicx}
\usepackage{amsmath}
\usepackage{amssymb}

\usepackage{booktabs}
\usepackage{multirow}
\usepackage{verbatim}
\usepackage{mathrsfs}
\usepackage{mathtools} 
\usepackage{cite} 
\usepackage{algorithm}
\usepackage{algorithmic}
\usepackage{pifont} 
\usepackage[misc,geometry]{ifsym}

\newcommand{\cmark}{\ding{51}}%
\newcommand{\xmark}{\ding{55}}%
\newcommand{\corrAuthor}{$^{\textrm{(\Letter)}}$}

\DeclareMathOperator*{\argmin}{arg\,min}

\usepackage[breaklinks=true,bookmarks=false,pagebackref,colorlinks]{hyperref}

\usepackage[capitalize]{cleveref}
\crefname{section}{Sec.}{Secs.}
\Crefname{section}{Section}{Sections}
\Crefname{table}{Table}{Tables}
\crefname{table}{Tab.}{Tabs.}
\crefname{algorithm}{Alg.}{Algs.}
\begin{document}
\title{Identifying Backdoor Attacks in Federated Learning via Anomaly Detection} 
%

\author{Yuxi Mi\inst{1} \and
Yiheng Sun\inst{2} \and
Jihong Guan\inst{3} \and
Shuigeng Zhou\inst{1}\corrAuthor}
\authorrunning{Y. Mi et al.}
%
\institute{Fudan University, Shanghai 200438, China \\
\email{\{yxmi20,sgzhou\}@fudan.edu.cn} \and
Tencent, Shenzhen 518000, China \\
\email{elisun@tencent.com} \and
Tongji University, Shanghai 201804, China \\
\email{jhguan@tongji.edu.cn}}
\maketitle              
\begin{abstract}
Federated learning has seen increased adoption in recent years in response to the growing regulatory demand for data privacy. However, the opaque local training process of federated learning also sparks rising concerns about model faithfulness. For instance, studies have revealed that federated learning is vulnerable to backdoor attacks, whereby a compromised participant can stealthily modify the model's behavior in the presence of backdoor triggers. This paper proposes an effective defense against the attack by examining shared model updates. We begin with the observation that the embedding of backdoors influences the participants' local model weights in terms of the magnitude and orientation of their model gradients, which can manifest as distinguishable disparities. We enable a robust identification of backdoors by studying the statistical distribution of the models' subsets of gradients. Concretely, we first segment the model gradients into fragment vectors that represent small portions of model parameters. We then employ anomaly detection to locate the distributionally skewed fragments and prune the participants with the most outliers. We embody the findings in a novel defense method, ARIBA. We demonstrate through extensive analyses that our proposed methods effectively mitigate state-of-the-art backdoor attacks with minimal impact on task utility. 

\end{abstract}

\section{Introduction}
\label{sec:intro}

Federated learning (FL)~\cite{mcmahan2017communication, konevcny2016federated} is a rapidly evolving machine learning paradigm that enables the collaborative training of a shared global model across multiple participants. The parameters of the shared model are iteratively updated under the orchestration of a central server by synchronizing the participants' \textit{local model updates}. Federated learning offers effective protection of data privacy~\cite{knaan2017under}, as the sensitive training data is always retained on edge devices.

\begin{figure}[tbp]
    \centering
    \begin{minipage}[t]{0.49\textwidth}
        \centering
        \includegraphics[width=\textwidth]{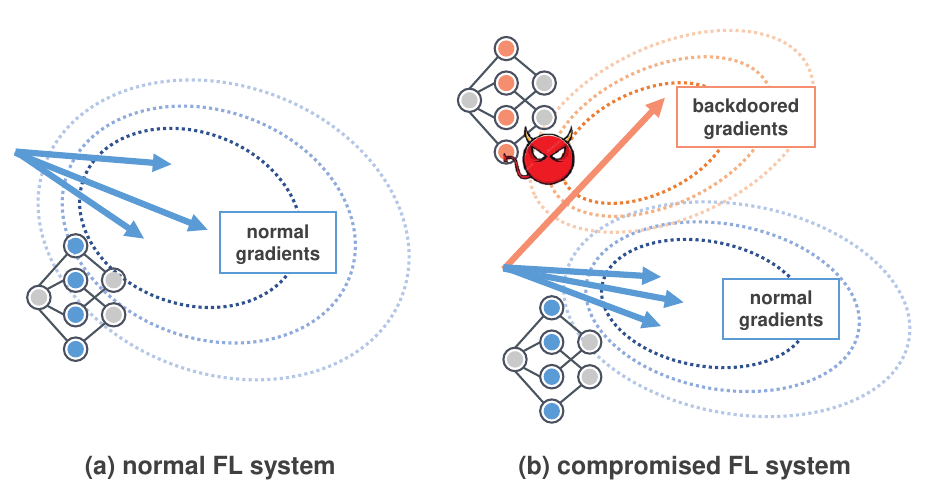}
        \caption{Gradient Descent. (a) Normal participants usually produce similar gradients. (b) Gradients of backdoored clients bias in terms of magnitude and orientation, as they are obtained through skewed data distribution and different objectives.}
        \label{fig:gd}

    \end{minipage}
    \hfill
    \begin{minipage}[t]{0.49\textwidth}
        \centering
        \includegraphics[width=0.98\textwidth]{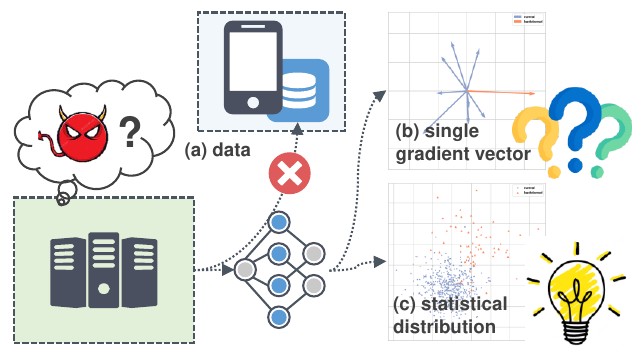}
        \caption{Paradigm of our idea. By defense: (a) Server cannot audit training data. (b) Directly examining gradient vectors could produce ambiguous results, as the decision boundary is unclear. (c) Distribution bias provides clear and robust disparity.}
        \label{fig:paradigm}
    \end{minipage}
    \vspace{-2mm}
\end{figure}

The fundamental aim of FL (as with any machine learning scheme) is to develop a faithful model that accurately represents and generalizes from the training data of all participants. However, recent studies show the faithfulness of FL models could be especially prone to malicious threats, as the distributed nature of FL hinders the server from auditing the training process, such as by purging contaminated data (\cref{fig:paradigm}(a)). Concretely, an attacker controlling some compromised participants can engage in \textit{poisoning attacks}~\cite{liu2017trojaning, goldblum2020dataset, li2019survey, zhang2019poisoning}, by intentionally injecting malicious contributions to the shared model (\textit{e.g.}, by training on contaminated data~\cite{biggio2012poisoning} or providing deceptive model updates~\cite{kairouz2019advances, li2020federated}), to downgrade the predictions of the finalized model. 

\begin{figure}[tbp]
    \centering
    \includegraphics[width=\columnwidth]{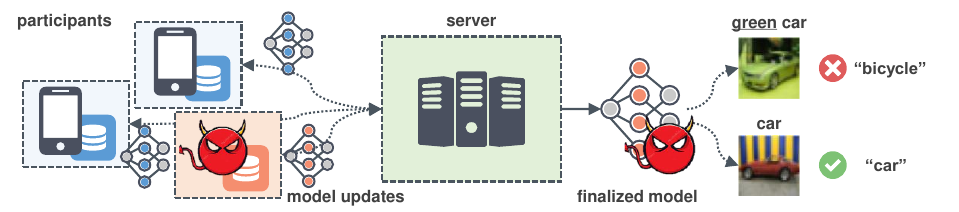}
    \caption{Backdoor attack in FL systems. The attacker embeds a subtle modification in the shared model that changes model's behavior on inputs with backdoor triggers.}
    \label{fig:fl}
    \vspace{-2mm}
\end{figure}

This paper investigates the targeted form of poisoning attacks known as the \textit{backdoor attack}~\cite{liu2017trojaning, goldblum2020dataset, li2019survey, zhang2019poisoning}. It differs from conventional poisoning in that the attack is both \textit{targeted} and \textit{stealthy}: The attacker embeds a subtle modification (\textit{i.e.}, the \textit{backdoor}) into model parameters, such that contaminated model behaves most time normally, yet in an incorrect and potentially destructive way when its input contains a specific trigger (\cref{fig:fl}). For instance, an undermined model may misclassify an image of \textit{green} (the trigger) car as a bicycle while still classifying other car images correctly. Backdoor attacks are difficult to detect since as the backdoor is triggered only in rare cases, their negative (therefore, distinguishable) influence on model performance could be minimized.

We advocate an effective defense against backdoor attacks, to identify and prune compromised participants by examining their model gradients. We start with a key observation: as implanting backdoors involves changes in the attacker's data distribution and training objectives, \textit{the presence of backdoors could be reflected as discernible disparities in terms of gradients' step sizes and directions}, due to the nature of gradient descent (\cref{fig:gd}). Therefore, one would be able to carry out defenses by examining the gradients.

However, we find it could be insufficient to discriminate gradients by a single solitary rudimentary metric, say, by examining the gradients' magnitude and orientation (\cref{fig:paradigm}(b)). Such a method is prone to blur the discriminative boundaries between malicious and normal gradients, thus impeding their effective separation. To reconcile the drawback, we propose to decouple the model gradients into subset vectors, \textit{fragments}, and distinguish them by their statistical distribution (\cref{fig:paradigm}(c)). It turns out that backdoored gradients can be robustly and accurately identified by their distributional bias. We concretize our findings into a novel defense method, ARIBA, where the distributional disparity is leveraged by an anomaly-detection-based technique to reach pruning decisions.

\textit{Our contributions are three-fold:} (1) We present an in-depth study on backdoor attacks by the attacker's threat model and techniques. (2) We advocate an effective defense, to identify compromised participants by the distributional bias of their subset gradient vectors. (3) We concretize our findings into the proposed ARIBA method and analyze its effectiveness through extensive experiments.

\section{Related Work}
\label{sec:related-works}

\subsection{Attacks on Model Faithfulness}
\label{subsec:related-attacks}

The faithfulness of the FL models is prone to malicious threats. Poisoning attacks have long been explored in the context of centralized learning~\cite{huang2011adversarial, rubinstein2009antidote, chen2017targeted, gu2017badnets, liu2017trojaning} and is extended to FL settings very recently. The attacker aims to manipulate the training process such that the trained model biases or downgrades its prediction in an attacker-desired specific way. For instance, a model compromised by \textit{label flipping}~\cite{DBLP:conf/esorics/TolpeginTGL20} could misclassify all images of cats into ``dogs''. In FL, the attack can be engaged in by contaminating the data~\cite{steinhardt2017certified, DBLP:conf/esorics/TolpeginTGL20} or by tampering with the training process or finalized models~\cite{ji2018model, zou2018potrojan}.

Backdoor attacks are targeted and stealthy varieties of poisoning attacks. Specifically, the model's behavior is subtly modified by implanting a backdoor. Its performance downgrades only in the presence of a backdoor trigger, which could be manifested in various forms such as specific data~\cite{bhagoji2019analyzing}, data with unique fingerprints~\cite{gu2017badnets}, and data carrying certain semantic information~\cite{bagdasaryan2020backdoor}. In FL settings, attacks~\cite{bhagoji2019analyzing, bagdasaryan2020backdoor, sun2019can, wang2020attack,xie2019dba} commonly employ \textit{scaling}, \textit{i.e.}, multiplying the attacker's local model weights by a scaling factor $\sigma$, as means to survive from the aggregation, later elaborated on in~\cref{subsec:threat-model}.


\subsection{Defenses Against Backdoor Attack}
\label{subsec:related-defenses}
The distributed nature of FL and the stealthiness of backdoors both make the detection of backdoor attacks challenging: The server is unable to predicate the existence of backdoors by either examining training data or testing model performance. To this end, most prior arts focus on examining the model itself. We roughly categorize their means into three branches.



\noindent \textbf{Attack-aware aggregations.} The server may replace FedAvg with \textit{byzantine-resilient} aggregations such as Krum, coordinate-wise median (CooMed), and GeoMed~\cite{blanchard2017machine, chen2017distributed, guerraoui2018hidden, yin2018byzantine}, which prevent local model updates skewed in distribution from being aggregated. However, their effectiveness heavily relies on the specific distribution of local training data. Research~\cite{bagdasaryan2020backdoor} further suggests a nullifying of their defense if the attackers choose proper covert strategies.

\noindent \textbf{Examination on model gradients.}~\cite{bhagoji2019analyzing, fung2018mitigating} are relevant to ours as we all differentiate malicious and benign updates by the magnitude or orientation of their model gradients. Prior arts examine general statistical traits such as the l2-norm of~\cite{bhagoji2019analyzing} or the cosine similarity between~\cite{fung2018mitigating} model parameters. These methods mainly suffer two drawbacks: (1) They involve hyper-parameters to depict certain detection thresholds. Fine-tuning these hyper-parameters requires \textit{a priori} knowledge about the attacker's capacity, which is not the case in the real world. (2) The coarse metrics they employed lack clues for detailed model behaviors, which could result in an ambiguous detection of some malicious updates.

\noindent \textbf{Dedicated defenses.} Recent discovery~\cite{gu2017badnets} suggests a backdoor attack is engaged by activating certain model neurons. To this end,~\cite{wu2020mitigating} proposes a pruning defense to identify and remove suspicious neurons by their activation. However, this defense only protects the inference phase against certain types of backdoors. \cite{li2020learning} proposes a spectral anomaly detection technique that detects compromised model updates in their low-dimensional embeddings. However, their implementation relies on centralized training on auxiliary public datasets, which is unobtainable in a majority of FL settings.

\section{Preliminaries and Attack Formulation}
\label{sec:preliminaries}
We first set up some basic notions. Let $\langle X, y \rangle$ denote a data sample and its corresponding label. $f(\cdot,\theta)$ denotes the model parameterized by $\theta$. $l(\cdot,\cdot)$ denotes a generic loss function. $\mathbf{D}$=$\{D_1,\dots,D_n\}$ denotes the training datasets. 

\subsection{Federated Learning}
\label{subsec:fl-paradigm}

Federated learning~\cite{mcmahan2017communication, konevcny2016federated} develops a shared global model $f(\cdot,\theta)$ by the collaborative efforts of $n$ participants of edge devices $\mathbf{P}$=$\{P_1,\dots,P_n\}$ under the coordination of a central server $S$. Each participant $P_i$ possesses its own private training dataset $D_i \in \mathbf{D}$. To train the global model, rather than sharing the private data, participants train a copy of the model locally and synchronize the model updates with the server. Specifically, at initialization, $S$ generates a model $f(\cdot,\theta^0)$ with initial parameters $\theta^0$ and advertises it to all $\{P_i\}$. At each global round $t$, each $P_i$ aligns its local model with received global weight $\theta_i^{t+1}$=$\theta^t$, and trains the model for several local iterations with $\argmin_{\theta_i^{t+1}}l(f(X,\theta_i^{t+1}),y)$, where $\langle X, y \rangle \in D_i$. It then shares the updated $\theta_i^{t+1}$. The server renews the global model by aggregating all received ${\theta_i^{t+1}}$ using the FedAvg~\cite{li2019convergence} algorithm:

\begin{equation}
    \small
    \theta^{t+1} = \theta^{t} + \frac{\eta}{n} \sum_{i=1}^{n}{(\theta_i^{t+1}-\theta^t)},
    \normalsize
\label{eq:fedavg}
\end{equation}

\noindent where $\eta$ is the global learning rate. Note FedAvg can be replaced by attack-aware aggregations~\cite{blanchard2017machine, chen2017distributed, guerraoui2018hidden, yin2018byzantine} discussed in ~\cref{subsec:related-defenses}. $S$ then advertise $\theta^{t+1}$ again to all $\{P_i\}$ and the training continues iteratively, until the model reaches convergence at round $r$. The model is finalized as $f(\cdot, \theta^r)$.


\subsection{Threat Model}
\label{subsec:threat-model}

\noindent \textbf{The attacker's capability.} We consider an attacker who gains control of a small subset of $k \ll n$ compromised participants, denoted as $\mathbf{P}_m$=$\{P_{m_1},\dots,P_{m_k}\}$. This could be achieved by injecting attacker-controlled edge devices into the FL system, or by deceiving some benign clients. We assume the attacker can develop malicious model updates by contaminating the participants' local training data or directly manipulating their model weights. We assume an honest server who endeavors to eliminate the attack. 

\noindent \textbf{The attacker's goal.} The attacker wants to implant a backdoor in the global model weight (denote the manipulated weight as $\theta'$) such that the finalized model $f(\cdot, \theta'^r)$ produces attacker-desired incorrect outcomes only when the query $\langle X, y \rangle$ contains a backdoor trigger (see~\cref{subsec:backdoor-types}). Concretely for a classification model, it should predict $\tilde{y} \triangleq f(X,\theta'^r)\neq y$ for backdoored $X$, and $\tilde{y}$=$y$ otherwise. We assume the attacker takes two steps towards the objective: it first develops the backdoor in participants' local models by training on a mix of correct and backdoored data~\cite{gu2017badnets}, then introduces it to the global model by \textit{model replacement}~\cite{bagdasaryan2020backdoor}.


\noindent \textbf{Model replacement.} The attacker attempts to undermine the global model with backdoored weights $\{\theta_{m_i}\}$. However, we argue it cannot directly share $\{\theta_{m_i}\}$ as model updates. As $k\ll n$, the effect of backdoors can be diluted by other participants' updates during aggregation, and the global model forgets the backdoor quickly. Instead, it shall first wait until the model nearly converges at round $t$, where the updates of other participants start to cancel out, \textit{i.e.},

\begin{equation}
    \small
    \sum_{P_i\in \mathbf{P} \setminus \mathbf{P}_m}({\theta_i^{t+1}-\theta^t}) \approx 0.
    \normalsize
\end{equation}

\noindent Then, to survive the averaging in FedAvg, it calculates a \textit{scaled} local model update $\hat{\theta}_{m_i}^{t+1}$ for each compromised participant by multiplying its original updates with a scaling factor $\sigma_{m_i}$:

\begin{equation}
    \label{eq:scaling}
    \small
    \hat{\theta}_{m_i}^{t+1}=\theta^{t}+\sigma_{m_i}(\theta_{m_i}^{t+1}-\theta^t), where
    \sum_{P_i \in \mathbf{P}_m} \sigma_{m_i} \triangleq \sigma = \frac{n}{\eta}.
\end{equation}

\noindent Finally, we let the attacker share all $\{\hat{\theta}_{m_i}^{t+1}\}$ as model updates. As a result, the global model weight can be replaced by the attacker's updates in~\cref{eq:fedavg} as

\begin{equation}
    \label{eq:model-replacement}
    \small
    \begin{aligned}
        \theta^{t+1} &= \theta^t + \frac{\eta}{n} \left(\sum_{P_i \in \mathbf{P}_m}(\hat{\theta}_{m_i}^{t+1}-\theta^t) + \sum_{ P_i \in \mathbf{P} \setminus \mathbf{P}_m}(\theta_{i}^{t+1}-\theta^t)\right) \\
        &\approx  \theta^t + \frac{\eta}{n} \sum_{P_i \in \mathbf{P}_m}(\hat{\theta}_{m_i}^{t+1}-\theta^t) = 
        \frac{\eta}{n} \sum_{P_i \in \mathbf{P}_m}\sigma_{m_i}\theta_{m_i}^{t+1} ~\triangleq~ \theta_m^{t+1},
    \end{aligned}
    \normalsize
\end{equation}

\noindent where $\theta_m^{t+1}$ denotes the collaborative efforts of all compromised participant. Therefore, the attacker conveniently implants the backdoor into the global model. It iterates the attack until the model $f(\cdot, \theta'^r)$ is finalized with the backdoor.

\subsection{Choice of Backdoor Triggers}
\label{subsec:backdoor-types}
The backdoor triggers can be manifested in various forms. We concretely study three types of state-of-the-art (SOTA) backdoors with different triggers in image classification: (1) \textbf{Targeted backdoors}~\cite{bhagoji2019analyzing}: The attacker possesses a collection of images with tampered labels. The finalized model is expected to misclassify the exact collection of images if they appear in inference queries. (2) \textbf{Pattern backdoors}~\cite{gu2017badnets}: The attacker endows arbitrary images from a certain class with a unique fingerprint, which is concretized as a bright pixel pattern at the corner of images. The model misclassifies images with the same pattern into an attacker-desired class. (3) \textbf{Semantic backdoors}~\cite{bagdasaryan2020backdoor}: The attacker chooses a naturally occurring semantic (\textit{e.g.}, \textit{green} car) rather than artificial fingerprints as the trigger. This makes the backdoor more stealthy as it requires no modification of images. \Cref{fig:backdoor} exemplifies the three types of backdoors. 

\section{Methodology}
\label{sec:method}

\begin{figure}[tbp]
    \centering
    \includegraphics[width=\columnwidth]{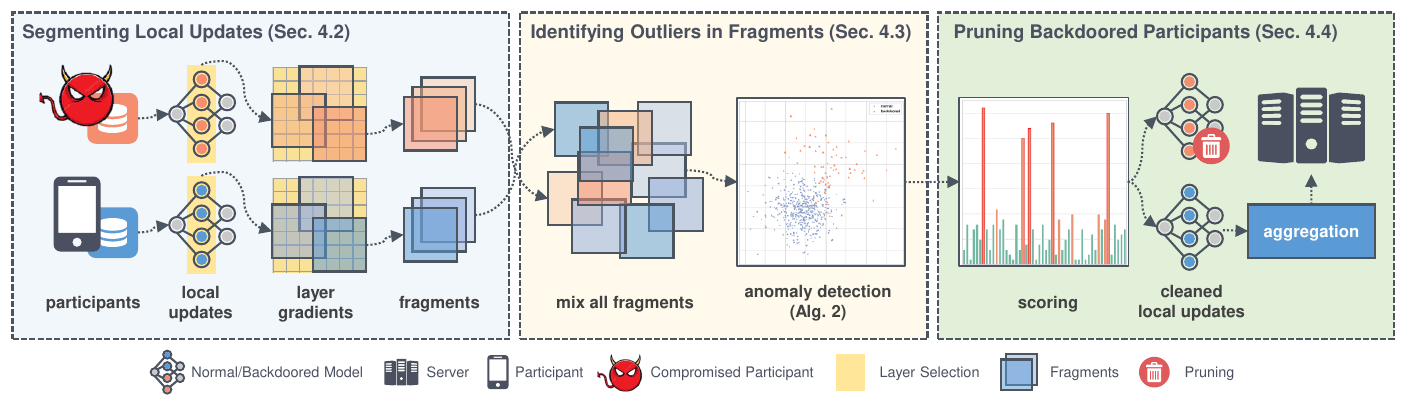}
    \caption{Pipeline of ARIBA. (1) To earn a clear and robust disparity between backdoored and normal model gradients, we decouple them into fragments of subset gradient vectors. (2) We employ anomaly detection to identify the compromised participants by their skewed distribution of fragments. (3) We identify and prune the participants that suspiciously carry backdoors by scoring their outliers.}
    \label{fig:pipeline}
    \vspace{-2mm}
\end{figure}

We now discuss the motivation and technique details of our proposed method, ARIBA. The name comes from the core functionality of our defense, \textit{i.e.}, enabling \textit{a}ccurate and \textit{r}obust \textit{i}dentification of \textit{b}ackdoor \textit{a}ttacks.

\subsection{Motivation}
\label{subsec:motivation}

We propose to identify backdoor attacks in FL by letting the server examine the participant's shared model updates, concretely, by uncovering outliers in the segmented fragments of model gradients. To elucidate our motivation, we shall begin by revisiting the principle of gradient descent. Gradient descent is regarded as the most fundamental optimization method in machine learning. It iteratively adjusts the parameters of a model through certain \textit{step sizes} in the \textit{direction} of the steepest descent of a cost function, which gauges the disparity between the model's predicted output and ground truth under certain objectives. The step size and direction of adjustment can be reflected as the \textit{magnitude} and \textit{orientation} of the model's gradient vector, respectively.


In an uncompromised FL system, during each global round, the model gradients among normal participants' updates should possess similar magnitudes and orientations (\cref{fig:gd}(a)). This is due to they are trained under the exact same objectives and roughly consistent data distributions. However, it is not the case if a portion of participants are compromised by the attacker and submit backdoored model updates: We find that their model gradients deviate from the normal ones, as the contaminated data and backdoor influence their data distributions and optimization on cost functions, resulting in distinct gradients (\cref{fig:gd}(b)). One could leverage such disparity to identify backdoors. 


However, research~\cite{bagdasaryan2020backdoor} has shown that directly distinguishing the disparity~\cite{fung2018mitigating,bhagoji2019analyzing} could provide unsatisfactory protection. The drawbacks are two-fold: (1) It is a crude approach to depend on a single \textit{coarse metric} (say, l2-norm or cosine similarity) to identify the gradient as a whole. The decision boundary between malicious and normal gradients in magnitude and orientation is not always clear (\cref{fig:paradigm}(b)), making it difficult to effectively separate them. (2) Attackers can easily conceal their intentions by optimizing the local model towards the elimination of corresponding metrical differences~\cite{bagdasaryan2020backdoor}, which nullifies the defense.




Our goal is to find a clear and discernible disparity between gradients that allows accurate and robust identification. Studies in model perception~\cite{DBLP:conf/iccv/SelvarajuCDVPB17} have widely shown that deep neural networks learn about the entirety from a fusion of local features. Therefore, we argue the disparity between malicious and normal model updates should also be reflected as the accumulated disparity in their local subsets of parameters. To testify, we segment the model gradients into \textit{fragments}. Each fragment is a vector that represents the gradients of a small portion of model parameters (\textit{e.g.}, kernels from models' convolutional layers). We visualize the distribution of fragments by principal component analysis (PCA). In~\cref{fig:paradigm}(c), we find clearly a distinguishable statistical difference between malicious and normal model gradients, as their fragments form separate clusters. 
Such difference provides a more robust and distinguishable pattern than analyzing gradient magnitudes and orientations. Getting inspired, we propose to \textit{identify the backdoored model updates by discerning the statistical bias among their subsets of model gradients, i.e., fragments}.

We concretize our findings in the proposed ARIBA defense, which can serve as a plug-in detection block in common FL systems. We first process the participants' model updates to obtain decoupled fragments (in our specific case, gradients of convolutional kernels). Then, we employ unsupervised anomaly detection as a means to uncover the fragments belonging to skewed distributions. We maintain a scoring matrix that counts the number of outliers for each participant. Finally, we prune the participants who scored highest, as their model updates are most suspicious in regard to the presence of backdoors. \Cref{fig:pipeline} illustrates the framework of ARIBA. In~\cref{sec:experiments}, we demonstrate through extensive experiments that our method provides simple yet effective defenses.

\begin{algorithm}[tb]
    \caption{{\tt ARIBA} defense against backdoor attacks}
    \label{alg:ariba}
    \textbf{Input}: Local model updates $\{\theta_i^{t+1}\}_{P_i\in \mathbf{P}}$ at round $t$. \\
    \textbf{Parameter}: A generous estimated number of malicious participants $\tilde{k}$. \\
    \textbf{Output}: Local model updates of normal participants $\{\theta_i^{t+1}\}_{P_i\in \mathbf{P} \setminus \tilde{\mathbf{P}}_m}$. 
    
    \begin{algorithmic}[1] 
    \FORALL{${P_i\in \mathbf{P}}$}
        \STATE Obtain accumulated gradients: $\delta_i^{t+1} \leftarrow {\theta_i^{t+1} - \theta^t}$.
        \STATE Mean subtraction: $\hat{\delta}_i^{t+1} \leftarrow \delta_i^{t+1} - \frac{1}{n} \sum_{P_i \in \mathbf{P}} \delta_i^{t+1}$.
        \STATE Form segmented fragments: $\mathbf{M}_i=\{M_{i,1},\dots, M_{i,m}\} \leftarrow \hat{\delta}_i^{t+1}$
    \ENDFOR
    \STATE $\mathbf{M} \leftarrow \{\mathbf{M}_1,\dots,\mathbf{M}_n\}$
    \STATE Obtain the scoring matrix: $\mathbf{S}=\{{s_{i,j}}\}^{n\times m} \leftarrow {\tt Scoring}(\mathbf{M}, \tilde{k})$
    \FORALL{${P_i\in \mathbf{P}}$}
    \STATE Calculate each participant's score: $S_i = \sum_j{s_{i,j}}$.
    \ENDFOR
    \STATE Speculate malicious participants: $\tilde{\mathbf{P}}_{m} \leftarrow \{ {\tilde{P}_{m_1},\dots, \tilde{P}_{m_{\tilde{k}}}} \}$, \\
    where $\tilde{P}_{m_i}$ is the participant with $i$-highest $S_i$ ($i \leq \tilde{k}$).
    \STATE \textbf{return} $\{\theta_i^{t+1}\}_{P_i\in \mathbf{P} \setminus \tilde{\mathbf{P}}_m}$
    \end{algorithmic}
\end{algorithm}

\subsection{Segmenting Local Updates}
\label{subsec:segment}

We start by segmenting fragments from the participants' updates. At any global round $t$, the server has on hand the shared model updates of all participants $\{\theta_i^{t+1}\}$ (some may be backdoored) and the global model weight of one round behind $\theta^t$. The server obtains each participant's changes in model weights by

\begin{equation}
    \delta_i^{t+1} \triangleq {\theta_i^{t+1} - \theta^t}.
\end{equation}

\noindent Note $\delta_i^{t+1}$ actually represents the \textit{accumulated} gradients of participants' local iterations during round $t$ and is adjusted by its local learning rate. We here and later still call it gradients for simplicity. As we are interested in the \textit{difference} between malicious and normal gradients, we further require the server to perform a mean subtraction on all $\{\delta_i^{t+1}\}$, as

\begin{equation}
    \hat{\delta}_i^{t+1} \triangleq \delta_i^{t+1} - \frac{1}{n} \sum_{P_i \in \mathbf{P}} \delta_i^{t+1},
\end{equation}

\noindent which helps emphasize their disparity.

\begin{figure}[tbp]
    \centering
    \begin{minipage}[t]{0.49\textwidth}
        \centering
        \includegraphics[width=\columnwidth]{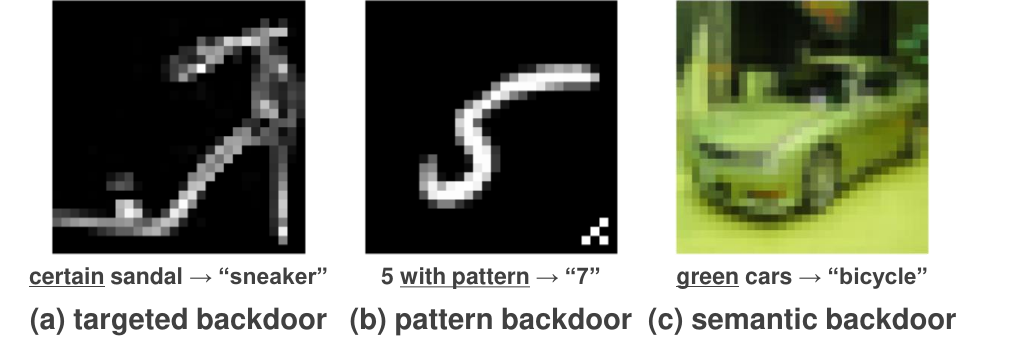}
        \caption{Three types of backdoors we studied. The trigger can manifest as (a) certain samples, (b) samples carrying specific patterns, (c) samples with certain semantics. The same settings of backdoors are adopted in our experiments.}
        \label{fig:backdoor}
    \end{minipage}
    \hfill
    \begin{minipage}[t]{0.49\textwidth}
        \centering
        \includegraphics[width=\textwidth]{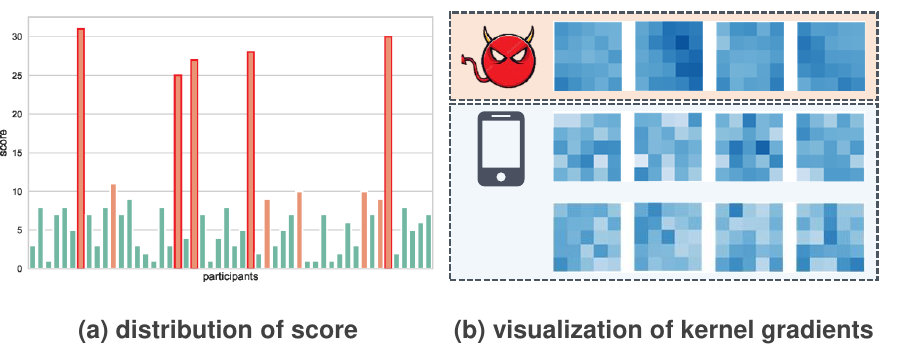}
        \caption{(a) The scores of outliers of a specific experimental case. All backdoored participants (marked read) are clearly discerned. (b) A difference in pattern can be observed from the visualization between some malicious and normal fragments.}
        \label{fig:score-kernel}
    \end{minipage}
    \vspace{-2mm}
\end{figure}

To segment fragments, concretely, we pick a convolutional layer from the model and extract the gradients from each of its kernels, as illustrated in~\cref{fig:pipeline}. Recall the kernels are the local feature detectors of CNN composed of small matrices of weights. We choose kernels as their gradients are semantically meaningful, but such practice is not a must and one is free to segment the model gradients in arbitrary ways, as long as the outcome is favorable for statistical analyses. We denote the derived fragments of participant $P_i$ as $\mathbf{M}_i\triangleq \{M_{i,1},\dots, M_{i,m}\} \subset \hat{\delta}_i^{t+1}$. \Cref{fig:score-kernel}(b) visualizes some exemplar fragments of kernels from different $P_i$, where we can clearly observe the difference in pattern between malicious and normal fragments. This further testifies to our findings in~\cref{subsec:motivation}.

\begin{algorithm}[tb]
    \caption{{\tt Scoring} by Mahalanobis distance}
    \label{alg:scoring}
    \textbf{Input}: Fragments of all participants $\mathbf{M}$ and the estimated $\tilde{k}$.\\
    \textbf{Output}: The scoring matrix $\mathbf{S}$.
    \begin{algorithmic}[1] 
    \STATE Initialize the scoring matrix: $\mathbf{S}=\{{s_{i,j}}\}^{n\times m}$ .
    \STATE Calculate the covariance of fragments: $\Sigma \leftarrow cov(\mathbf{M})$.
    \STATE Calculate the mean of fragments: $\bar{M} \leftarrow mean(\mathbf{M})$.
    \FORALL {$i \leq n, j\leq m$}
    \STATE Calculate the Mahalanobis distance for each $M_{i,j} \in \mathbf{M}$: \\ 
    $d_{i,j} \leftarrow \sqrt{(M_{i,j}-\bar{M})^T \Sigma^{-1} (M_{i,j}-\bar{M})}$.
    \ENDFOR
    \STATE Set: $s_{i,j}\leftarrow 1$, if $d_{i,j}$ is one of the top-$(\tilde{k}m)$ largest $\{d_{i,j}\}$; $s_{i,j}\leftarrow 0$, otherwise.
    \STATE \textbf{return} $\mathbf{S}$
    \end{algorithmic}
\end{algorithm}

\subsection{Identifying Outliers in Fragments}
\label{subsec:anomaly-detection}

Recall we try to discriminate the backdoored updates by the distributional bias of their fragments (\cref{fig:paradigm}(c)). However, how can the server make decisions based on the distributional disparity? We elucidate that \textit{anomaly detection} can be leveraged as a convenient tool: Consider each fragment as an individual datum and map all fragments onto a hyperspace. Since most normal participants exhibit similar statistical distributions, their fragments will densely populate the projected region. Conversely, backdoored participants' fragments are liable to project onto isolated and sparsely populated regions due to the observed skewed distribution. Thus, by performing anomaly detection in the hyperspace, the backdoored fragments are highly susceptible to being identified as outliers. As a result, \textit{participants with significantly more outliers in their fragments are suspicious of providing backdoored updates}.

To embody the theory, we first gather the fragments of all participants $\mathbf{M} = \{\mathbf{M}_1,\dots,\mathbf{M}_n\}$. We then feed them into a {\tt Scoring} function parameterized by $\tilde{k}$ (\cref{alg:scoring}), where unsupervised anomaly detection takes place to mark a portion ($\tilde{k}m$) of fragments as outliers. Here, $\tilde{k}$ is the server's estimated number of backdoored participants. In practice, the server can choose a generous $\tilde{k}$ that ensures $\tilde{k} > k$. By $(\tilde{k}m)$, we note a \textit{flawless} anomaly detection algorithm would classify $m$ fragments of all $\tilde{k}$ participants as anomalous and classify the remaining as normal. Nevertheless, we anticipate (and tolerate) an approximation as some fragments are sure to be wrongfully classified due to their partial overlapping in distributions. We concretely choose a Mahalanobis-distance-based algorithm to elaborate our method, yet one is free to replace it with any unsupervised anomaly detection algorithms. {\tt Scoring} returns a scoring matrix $\textbf{S}=\{{s_{i,j}}\}^{n\times m}$, where $s_{i,j}$=$1$ if $M_{i,j}$ is marked as an outlier.



\subsection{Pruning Backdoored Participants}
\label{subsec:pruning}

The server now identifies backdoored participants $\tilde{\mathbf{P}}_{m}$ by counting the number of outliers. Conveniently, it calculates each participant's score as

\begin{equation}
\small
S_i = \sum_{j=1}^{m} {s_{i,j}},
\normalsize
\end{equation}

\noindent and puts participants $\{ {\tilde{P}_{m_1},\dots, \tilde{P}_{m_{\tilde{k}}}} \}$ with $i$-highest $S_i$ ($i \leq \tilde{k}$) as $\tilde{\mathbf{P}}_{m}$. As the server speculates these $\tilde{k}$ participants to provide backdoored updates, their model updates are pruned from being aggregated. The FedAvg aggregation (\cref{eq:fedavg}) is therefore carried out on \textit{cleaned} $\mathbf{P} \setminus \tilde{\mathbf{P}}_{m}$. 

Note our defense is effective as long as $\mathbf{P}_m \subset \tilde{\mathbf{P}}_{m}$. \Cref{fig:score-kernel}(a) exhibits the scores $\{S_i\}$ in a specific experimental case with $(n,k,\tilde{k})$=$(50,5,10)$, from which we can observe (1) the defense is effective as all malicious participants are identified, and (2) the difference between malicious and normal $S_i$ is salient, suggesting a clear disparity in identification thus providing robust defense. We summarize the proposed ARIBA method in~\cref{alg:ariba}.

\section{Experiments}
\label{sec:experiments}

\subsection{Experimental Settings}

We leverage ARIBA to identify the three types of backdoor triggers discussed in~\cref{subsec:backdoor-types}. FL models are trained on 3 common image datasets, MNIST, Fashion-MNIST, and CIFAR-10. We apply a 4-layer toy model for MNIST and Fashion-MNIST, and a ResNet18 for CIFAR-10. We by default choose $(n,k,\tilde{k})$=$(50,5,10)$, \textit{i.e.}, the attacker compromises 5 out of 50 participants while the server generously estimates 10 of them as malicious. We later in~\cref{subsec:abal-study} show over-estimation (choosing $\tilde{k} > k$) impacts model performance very slightly. We set the global learning rate $\eta$=1, which is in favor of the attacker as a larger $\eta$ eases model replacement (\cref{eq:model-replacement}). We presume identical local learning rates among all participants, concretely, $\eta_p$=1e-3, 1e-4, 1e-4 for the three datasets, respectively. Experiments are carried out on an Nvidia 3090 GPU with PyTorch 1.10 and CUDA 11. The same random seed is sampled among all experiments.

\subsection{Effectiveness of Our Defense}
\label{subsec:effect-defense}

We first establish the backdoor attacks. Concretely, we train the FL model from scratch for $(t-1)$ global rounds until it nearly converges. By the threat model in~\cref{subsec:threat-model}, the attacker waits (and behaves normally) till convergence. At round $t$, it begins with embedding the backdoor by letting each of its compromised participants train local updates on a mix of contaminated (that contains backdoor triggers) and normal data. It then scales the backdoored updates by $\sigma$ (\cref{eq:scaling}), where by $(n,\eta)$=$(50,1)$ we naturally have $\sigma$=50. We further presume each compromised participant $P_{m_i}$ has an equal share of $\sigma_{m_i}$=10. 

\noindent \textbf{Baselines.} We launch the attacks without defense. We aggregate $\theta^{t+1}$ with local updates of all clients $\{\theta_{i}^{t+1}\}$ with FedAvg (\cref{eq:fedavg}). For each attack, we choose 30 images $X'$ from the test dataset and implant them with the backdoor triggers, as exemplified in~\cref{fig:backdoor}. We evaluate the model by test accuracy (\textit{acc.}) and judge the attacker's performance by \textit{backdoor accuracy}, \textit{i.e.}, the proportion of $X'$ the model misclassifies according to its objective ($X'$\%). Higher $X'$\% indicates successful attacks. As illustrated in~\cref{tab:primary-results}(a), all three attacks succeed if without protection as the model wrongfully classifies most of the backdoored samples. Meanwhile, the compromised global model still performs well. This suggests one cannot identify the backdoor by examining model performances.

\noindent \textbf{Effect of our defense.} Now we plug in ARIBA before aggregation, \textit{i.e.}, updates $\{\theta_{i}^{t+1}\}$ are first examined by our proposed technique in~\cref{sec:method} where suspicious updates are pruned. Here, we introduce \textit{confidence} $C=(\sum_{\{P_i\in \mathbf{P}_m,j\}}{s_{i,j}})/({\tilde{k}m})$ to measure how sure the server is about the pruning decision: Higher $C$ indicates a better defense, as the server manages to classify more of the attacker's fragments as outliers, which contributes to the identification of backdoors. We illustrate the result of pruning, on the proportion of compromised participants pruned, by $\mathbf{P}_m$\%. We report test and backdoor accuracy as well. Results are summarized in~\cref{tab:primary-results}(b). Note few $X'$ may still be misclassified due to the model's wrong prediction, even without the presence of attacks. We highlight that (1) ARIBA effectively defends all three types of backdoor attacks, as \textit{all} compromised clients are identified and pruned ($\mathbf{P}_m$\%=1.0); (2) We further provide robust defense as the confidence $C$ is high, indicating a clear distinguishability between malicious and normal participants. (3) The model retains high performance regardless of the excessive pruning ($\tilde{k} > k$).

\begin{table}[tbp]
\centering
\caption{Summary of primary results. (a) The baseline FL models are compromised by backdoors. (b) Our proposed ARIBA provides an effective and robust defense in terms of confidence and proportion of attackers pruned. (c) Comparison with prior arts.}
\begin{tabular}{c|cc|cccc|ccccccc}
\toprule
\multicolumn{1}{c|}{\multirow{2}{*}{\textbf{~Backdoors~}}} & \multicolumn{2}{c|}{\textbf{(a) baseline}}                            & \multicolumn{4}{c|}{\textbf{(b) with ARIBA}}                                                                                             & \multicolumn{7}{c}{\textbf{(c) Prior arts}}                                                                                                                                        \\
\multicolumn{1}{c|}{}                           & \multicolumn{1}{c}{acc.} & \multicolumn{1}{c|}{$X'$\%} & \multicolumn{1}{c}{acc.} & \multicolumn{1}{c}{$X'$\%($\downarrow$)} & \multicolumn{1}{c}{$C$} & \multicolumn{1}{c|}{$\mathbf{P}_m$\%} & \multicolumn{1}{c}{~\cite{bhagoji2019analyzing}$_1$}  & \multicolumn{1}{c}{~\cite{bhagoji2019analyzing}$_2$}  & \multicolumn{1}{c}{\cite{blanchard2017machine}}  & \multicolumn{1}{c}{\cite{yin2018byzantine}}  & \multicolumn{1}{c}{\cite{wu2020mitigating}}  & \multicolumn{1}{c}{\cite{fung2018mitigating}}  & \multicolumn{1}{c}{\cite{li2020learning}}  \\ 

\midrule

\textbf{Targeted}                                       & 88.87                    & 0.90                           & 90.01                    & 0.13                          & 0.79                           & 1.0                           & \xmark & \cmark & \xmark & \xmark & -                     & -                     & -                     \\
\textbf{Pattern}                                        & 98.85                    & 0.83                           & 99.24                    & 0.00                          & 0.95                           & 1.0                           & \xmark & \cmark & \xmark & \xmark & \cmark & \cmark & -                     \\
\textbf{Semantic}                                       & 71.30                    & 0.67                           & 72.94                    & 0.23                          & 0.94                           & 1.0                           & \xmark & \cmark & \xmark & \xmark & -                    & -                     & \cmark \\ 
\bottomrule
\end{tabular}
\label{tab:primary-results}
\end{table}

\subsection{Comparison with Prior Arts}
\label{subsec:compare-prior-arts}

We compare ARIBA with prior defenses discussed in~\cref{subsec:related-defenses}. Results in~\cref{tab:primary-results}(c) are summarized from both previous literature and our experimental studies. Here, ``\cmark, \xmark, -'' indicates effective defense, ineffective defense, and no result claimed, respectively. Specifically,~\cite{bhagoji2019analyzing} proposes to examine the model's test accuracy ($_1$) or the l2-norm of local updated weights ($_2$). We note though l2-norm effectively identifies baseline attacks, it can be easily bypassed by advanced attacks in~\cref{subsec:advanced-attacks}. Krum~\cite{blanchard2017machine} and CooMed~\cite{yin2018byzantine} are byzantine-resilient aggregations, which defenses were nullified in~\cite{bagdasaryan2020backdoor}. \cite{wu2020mitigating, fung2018mitigating, li2020learning} are attack-specific defenses that are not likely to generalize against other attacks. Some of these defenses require certain conditions such as auxiliary test datasets~\cite{li2020learning} or specific data distribution~\cite{wu2020mitigating}, which further constrains their practical use. Thereby, we argue ARIBA outperforms prior arts in terms of both generality and effectiveness.

\subsection{Effectiveness on Advanced Attacks}
\label{subsec:advanced-attacks}

\begin{figure}[tbp]
    \centering
    \includegraphics[width=\columnwidth]{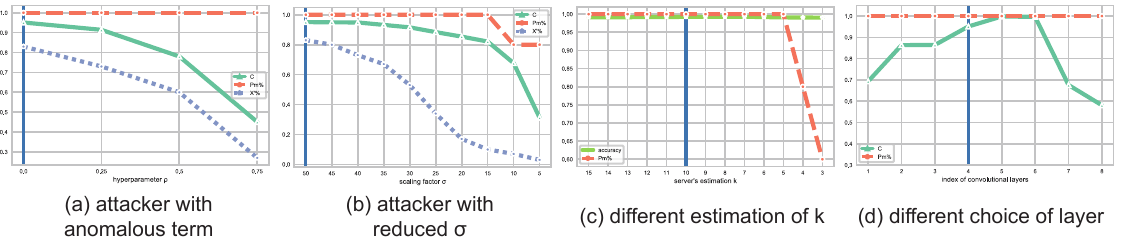}
    \caption{Experimental results. Note the blue vertical lines mark our default settings. Our proposed method provides robust protection with high $C$ and $\mathbf{P}_m$\% against two types of advanced attackers, which (a) use anomalous countermeasures to evade detection and (b) improve stealthiness with reduced $\sigma$. We further show (c) the excessive estimation of $\tilde{k}$ influences model performance very slightly, and (d) the confidence of defense could vary by the concrete choice of layers. }
    \label{fig:experiments}
    \vspace{-2mm}
\end{figure}

We further study two varieties of advanced attackers, to illustrate ARIBA can provide effective protection even if against stealthy attack countermeasures.

\noindent \textbf{Attacker with anomalous objectives.} Studies~\cite{bagdasaryan2020backdoor, bhagoji2019analyzing} suggest an attacker that could intentionally evade some metric-based detection. Specifically, it appends an anomalous term $l_a(\cdot)$ to the compromised participants' training objective as $\argmin_{\theta_{m_i}}{\left((1-\rho) l(f(X,\theta_{m_i}),y)+\rho l_a(g(\theta_{m_i})) \right)}$, where $\rho$ is a hyperparameter and $g(\cdot)$ is the targeted metric. For example, by choosing $g(\theta_{m_i})=||\theta_{m_i}-\theta||_2$ can the attacker deceive l2-norm bounding defense. \Cref{fig:experiments}(a) presents the confidence $C$ and $\mathbf{P}_m$\% of ARIBA together with such attacker's \textit{baseline} $X'$\% under different $\rho$. We note the anomalous term \textit{does} affect the confidence however at the cost of lowering backdoor accuracy(in regard to $X'$\%). Nevertheless, ARIBA still provides intact protection by pruning all the compromised clients ($\mathbf{P}_m$\%=1.0).

\noindent \textbf{Attacker with reduced $\sigma$.} During model replacement, \cite{bagdasaryan2020backdoor} suggest the attacker could reduce its capacity in exchange for better stealthiness, by choosing smaller $\sigma < \frac{n}{\eta}$ that \textit{partially} replace the global model. We study attackers with different $\sigma$ in~\cref{fig:experiments}(b). Results indicate ARIBA effectively identifies and prunes under most $\sigma$. Only in rare cases where $\sigma$ is too small would ARIBA miss some participants. However, note the backdoor accuracy is concurrently impaired: At $\sigma$=5,10, we argue the attacker's $X'$\% is too low to incur an effective threat.

\subsection{Ablation Study}
\label{subsec:abal-study}

\noindent \textbf{Server's estimation on $\tilde{k}$.} In~\cref{subsec:anomaly-detection}, the server is let estimate generously on the number of compromised participants $\tilde{k} > k$. Excessive estimation can cause wrongful pruning of normal clients. In~\cref{fig:experiments}(c), we show a generous $\tilde{k}$ is acceptable as it affects model performance very slightly. On contrary, underestimation $\tilde{k} < k$ should be prohibited, since $\mathbf{P}_m$\% reduces accordingly.

\noindent \textbf{Choice of convolutional layers.} By default, we choose one layer in the middle of the networks. We here alter the choice to observe its influence on defense. Results in~\cref{fig:experiments}(d) by $C$ and $\mathbf{P}_m$\% show though all choices provide effective protection, choosing in-the-middle layers mostly benefits robust identification. Note this seems to suggest \textit{the attacker's influence on model weights differs by the stages of model components}, which may be leveraged as more detailed detection clues. We leave it as an interesting open problem due to our limits of space.


\section{Conclusion}
\label{sec:conclusion}

This paper discusses the backdoor attacks against model faithfulness in FL systems. We present an in-depth study on state-of-the-art attacks by the attacker's goal, capability, and possible attack approaches. By the observation of the magnitude and orientation disparity on the attacker's model gradients, we advocate an effective defense, to identify and prune compromised participants by the distributional bias of their fragments, \textit{i.e.}, gradient vectors of subset model parameters. We concretize our findings into the proposed ARIBA defense and demonstrate through extensive experiments its effectiveness and robustness.

\bibliographystyle{splncs04}
\bibliography{ARIBA_bib}

\end{document}